# SOCIAL DISTANCE DETECTION USING DEEP LEARNING AND RISK MANAGEMENT SYSTEM

Jaya Aravindh V. V   jayaaravindh.vv2019@vitstudent.ac.in

Dr. Sangeetha. R. G. sangeetha.rg@vit.ac.in

Vellore Institute of Technology, Chennai Campus

---

## Abstract

An outbreak of the coronavirus disease which occurred three years later and it has hit the world again with many evolutions. The effects on the human race have already been profound. We can only safeguard ourselves against this pandemic by mandating a "Face Mask" also maintaining the "Social Distancing." The necessity of protective face masks in all gatherings is required by many civil institutions in India. As a result of the substantial human resource utilization, personally examining the whole country with a huge population like India, to determine whether the execution of mask wearing and social distance maintained is unfeasible. The COVID-19 Social Distancing Detector System is a single-stage detector that employs deep learning to integrate high-end semantic data to a CNN module in order to maintain social distances and simultaneously monitor violations within a specified region. By deploying current Security footages, CCTV cameras, and computer vision (CV), it will also be able to identify those who are experiencing the calamity of social separation. Providing tools for safety and security, this technology disposes the need for a labor-force based surveillance system, yet a manual governing body is still required to monitor, track, and inform on the violations that are committed. Any sort of infrastructure, including universities, hospitals, offices of the government, schools, and building sites, can employ the technology. Therefore, the risk management system created to report and analyze video streams along with the social distance detector system might help to ensure our protection and security as well as the security of our loved ones. Furthermore, we will discuss about deployment and improvement of the project overall.

**Acronyms:** YOLO-You Only Look Once, COCO-Common Objects in Context, DNN- Deep Neural Network

# I. Introduction

Crowds, or groups of people, are a common occurrence in human life. Everyday activities such as using the city's internal transportation system to work or visiting malls in retail environments, as well as social gatherings like college events, classes, and restaurants, as well as entertainment that includes theatres, public parks, Schools, Colleges are all some of the crowded places in which the public as a whole tends to be concentrated. Crowd control had become increasingly necessary and required in public areas for gatherings. This management is carried out by a close collaboration between the organizers and managers of the event, the emergency response teams, the local tending, the institutional authorities, the transportation authorities, the supervisors, and representatives of the public. The ability to effectively manage crowds depends on having a thorough grasp of participant spatial behaviors, psychological states, and interactions. It provides security by effectively addressing the circumstances that might be fueling the development of disasters as a result.

In the absence of centralized guidance, improperly coordinated behaviours could result in herding attitudes, according to social learning strategists who study crowd dynamics in humans. Wearing of masks is the suitable solution in order to stop the spread of infectious aerosols and droplets. The SARS Covid virus spreads from one's nose and mouth's secretions, when an infected person coughs (or sneezes), which could have adverse impacts on other individuals in crowded spaces. However, in a virus-infected environment, even those wearing masks may become infected. By wearing masks in crowds, people protects themselves and not contaminate it to others. So, for the sake to stop the contamination of the virus, we need an reliable surveillance steps that keeps an eye on people in the public domain. Monitoring social distance manually is quite tedious since it is challenging to observe continually with the human eye. Therefore, this is carried out automatically by our software program, which also highlights such individuals with red colour boxes when they violate social distance norms while monitoring people. CCTV cameras are used to automatically monitor people. The creation of this instrument is necessary to aid citizens and the government in locating and warning those responsible for an epidemic's rapid spread.

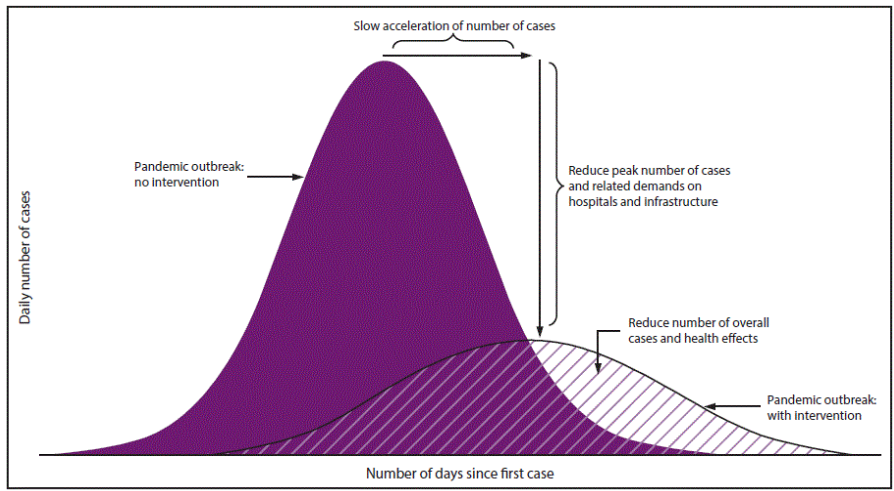

The Effect of Social distancing. As social distancing reduces and delayed the peak

This project suggests a model that makes use of the CNN method and was trained using the COCO Dataset. Outcomes are achieved by uploading footages of the people, to be identified from groups of people wearing masks and people who doesn't follow. Using dashboards with metrics, helps management maintain the crowd under control over the long term by detecting those who violate social distance in crowded places.

## II. Related Work

In recent times, artificial intelligence (AI) has been employed in numerous industries ranging from healthcare, Scientific Research, Space Exploration and entertainment media. One of the most remarkable features of deep learning is its ability to extract valuable insights and provide useful findings from data that can be used for decision-making and predictions. These unique characteristics of deep learning have led to more intelligent automation, reduced operational costs, and increased evolutionary capacity. However, despite the different approaches employed in various projects, each perspective has its limitations or achieves better results than others. From the methods used to feed images or videos, to the detection algorithms and optimization parameters utilized, everything varies depending on the project.

The paper from Z. Shao, G. Cheng, the use of an Unmanned Aerial Vehicle as a non-static platform for capturing crowd images has become an attractive option for social distance monitoring. These authors have showed a lightweight human detection network that employs human head detection to accurately identify pedestrians in real-time and calculate the distances between them based on UAV's feed. To improve the detection of objects like human heads, the researchers utilized PeleeNet Framework and incorporated multi-scale feature extraction and spatial calibration. Their experiments on the Merge-Head dataset showed that their method achieved an average precision of 92.22% and a frame rate of 76 FPS, which enables real-time detection in practical uses. Furthermore, the authors conducted experiments that revealed that multiscale features and spatial arrangement calibration significantly enhanced the results of the project.

In Dushyant Kumar Research. He has developed CNN-based techniques to detect the presence of humans in their work. They trained their model using the INRIA picture dataset and divided it into two modules for training and testing, respectively. The training module consisted of 2316 positive and 4046 negative images, while the testing module consisted of 100 positive and 100 negative images. The dataset comprised a total of 6562 images, of which 2416 were positive and 4146 were negative. During testing with real-time video sequences, the sliding window-based modules utilized a minimum window size of (64, 128), a step size of 10 cross each, and a downscale of 1.25. For human detection, the proposed technique employed a CNN architecture, while the sliding window approach was used for region proposal. The output from the human detector was used to calculate the distances between each pair of individuals observed. The social distancing algorithm marked people in red when they got too close to the permitted limit. This particular article is of significant importance to my project as it proposes a deep learning-based framework for automate the task of human detection, monitoring social distance from surveillance video. The project employs the YOLO v3 object detection model to detect, segregate pedestrians from the background and Deepsorting approach to track the detected people after implying bounding boxes. This study also compares the outcomes of the YOLO v3 model with other object detection models, including the faster region-based CNN (F-CNN) , single-shot detector (SSD), in terms of mean average precision (mAP), frames per second (FPS), and other important parameters related to object detection and classification. These results of the study show that the proposed framework has a balanced FPS score, which is better than other models.

Over the research, it was found that this paper which deals with different design and implementation was done and fairly good to be studied. The suggested mechanism employs an unmanned aerial vehicle (UAV), which operates autonomously to conduct the examination procedure. Initially, the UAV is fabricated by considering certain criteria such as the selection of components, determination of payload capacity, and subsequent assembly of the drone's constituent parts. Furthermore, the UAV is connected to a mission planning software program to calibrate the UAV's stability. In addition, a trained yolov3 algorithm with customized data sets is integrated into the drone's camera system. The camera system is configured to execute the yolov3 algorithm and identify whether individuals are maintaining social distancing protocols and adhering to mask-wearing mandates in public. This process is entirely automated and is performed by the drone without human intervention. In another piece of investigation, a deep learning model is provided to identify those who are not wearing masks and those who fail to keep a safe distance in order to contain the virus. People who violate the SOP are also recorded. The Single Shot Detector (SSD) is used in recommended model to extract feature elements, and Spatial Pyramid Pooling is then used to incorporate the retrieved features and increase the model's distinguishing abilities. In order to do the real-time social distance detection in embedded devices, the classifier uses the MobilenetV2 architecture as a framework, which makes the model extremely spontaneous, rapid, and efficient.

## III. Proposed System

The proposed system, Social Distance detector and its management system is developed by several components including Computer Vision methods (CV), Deep learning structure, Python's web application "Flask" framework is deployed in this development of the project.

1. Initially, for detection of the subject, which is humans in this case in the video's frame or image YOLOv3 object detection algorithm is used. We pass the video stream by frames

2. We detect "Humans" class from the rest by YOLO object detection, which does it from the features learnt by a convolutional neural network for object detection.

3. Then the bounding boxes are deployed and mapped to the human and the pairwise distances are calculated over the centroids of the rectangular boxes drawn on the people

4. Check if the pixel distance between any adjacent persons or boxes are less than the threshold pixels value.

5. Finally, count the number of violations and store in the local database for the flask framework to create the dashboard.

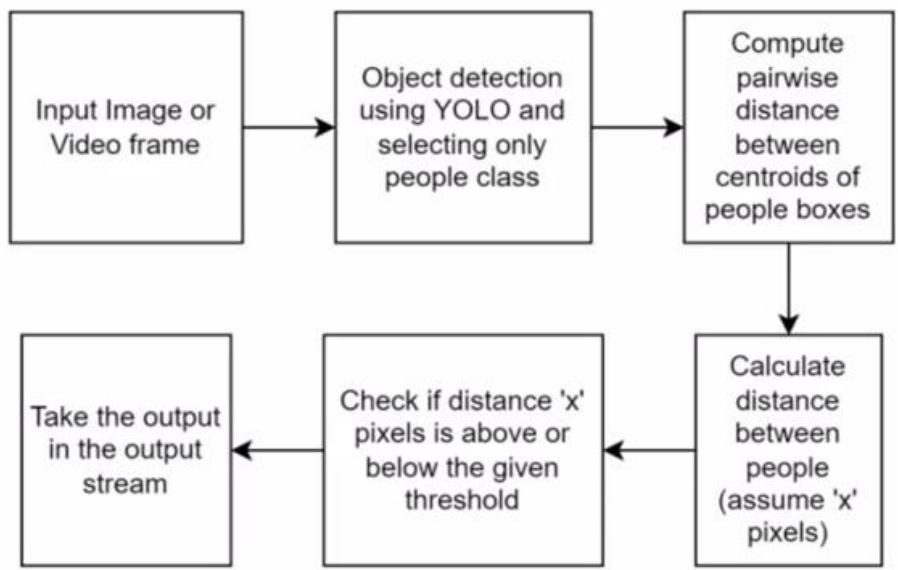

Workflow of the proposed system

## IV. Methodology

This section of the paper discusses the Design methodology of the project in a detailed manner of the necessities and working of the project.

Moreover, the project's backbone is YOLOv3, which runs on pre-trained with COCO-Dataset. The project works on Deep learning features as in "Image's Parts" with Non-Maxima Suppression and Darknet53. The image's parts or blob feature is used for the pre-processing of the image, and its processed with NMS for damping of the feeble or poor detections and Darknet53 as a framework for object detection with YOLO workflows. The Object detection model is tested and trained on COCO dataset, which contains 80 different types of classes with 33000 images, out of all persons is the only class we are focused. In this project, we run the project the pre-defined threshold distance to 50 pixels, the threshold at 0.3, and the configurations to 0.3 minimum. OpenCv, imutils, scipy, NumPy, and argparse are some of the libraries utilised in the project.

### A. Input Frames

To begin with, the initial step involves feeding the model with either image or video data obtained from the closed-circuit television (CCTV) system. The data is then segmented into individual frames for processing. Typically, the camera is positioned at a fixed angle, and the frames are converted to a 2D bird eye view to improve depth perception. The new view's dimensions are 448 pixels on all sides. The camera tuning process is carried out with OpenCV, and view's transformation is achieved through a measuring function that picks four points within the i/p image's frame or video frame, then maps them to the sides of a rectangular 2D image frame. After. It yields results a levelled frame, which then transformed into a horizontal plane by selecting the four endpoints of the frame and transforming them into a bird's eye view. This transformation facilitates easier distance calculations between individuals.

### B. Object Detection and Tracking

The Convolutional Neural Networks (CNN) framework is an efficient, also straightforward method to recognize objects. This model focuses solely on identifying areas belonging to the "Person" class, while disregarding sub-regions that are unlikely to have any objects. The formulation of extracting regions that contain our subjects is referred to as Region Proposals. These regions may differ in size and over bound with other regions. To address this issue, the Intersection Over Union (IOU) score is used, and the maximum non-suppression technique is applied to eliminate bounding boxes that overlap with each other. By utilizing a single regression problem to identify objects, this model's object detection technique effectively addresses computational complexity issues.

The You Only Look Once (YOLO) model is an advanced deep learning-based technique utilized for identifying objects. This method is particularly useful for real-time uses since it provides spontaneous and more precise results. YOLOv3 model is an object detection model capable of simultaneously learning and mapping the coordinates of the bounding box (t(x,y,w,h)), associated probabilities of classes labels, and object's confidence from both images and videos. Moreover, the YOLOv3 model has been pre-trained on the COCO dataset, which contains 80 distinct labels, including the "human" class.

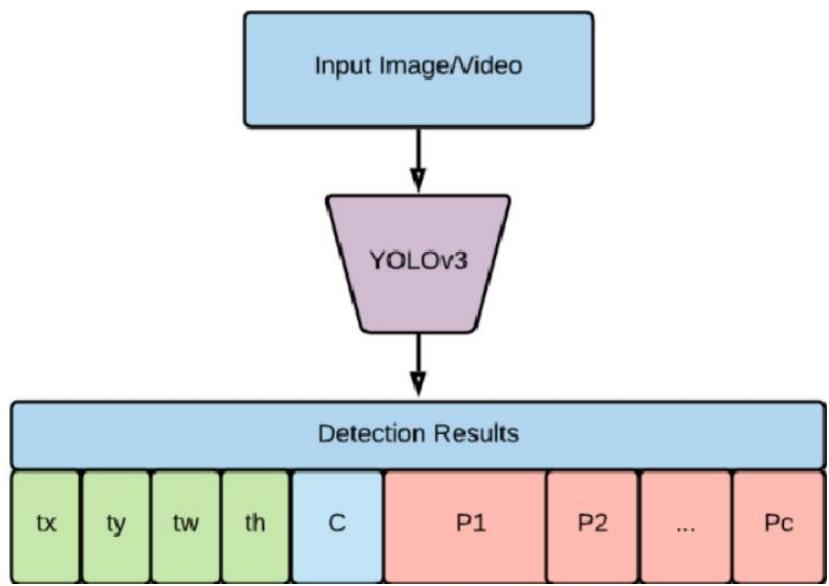

Detection of Pedestrians using YOLOv3

During preparation, a probability score is used. YOLOv3 predicts an definition-less score for each boxes of the sub-regions utilizing repulsive counters. That is, the item is predicted in a single run by simultaneously predicting several class probabilities and bounding boxes. To guarantee that the object recognition algorithm identifies each object only once and eliminates any inaccurate detections, the non-maximum suppression technique is utilized before returning the recognized objects and their associated bounding boxes.

## C. Distance Calculation

After recognising persons in the frame supplied as input, the model will detect the person and draw various bounding boxes on the person. As a result, several boxes may be created around the individual, which may be prevented by employing an algorithm known as Non-maximum suppression (NMS). NMS will consider the box with the best probability of covering that individual, and so just one box will be drawn around the person.

1. After the bounding boxes are drawn, we find centroid of the bounding boxes.

2. After finding centroid, we find distance between two centroids which will estimate the distance between the bounding boxes.

   To find the distance between two co-ordinates, the Euclidean distance estimation is utilized. Following this, the focus coordinates are converted into a square-shaped array. Next, the function filters the recognitions and separates them by individual class. The "convert back" function takes in parameters such as midpoint of the bounding box (x, y, w, h) of the bounding box. This function transforms the centre coordinates into rectangle coordinates and reverts back the converted points as x1, y1, x2, and y2.

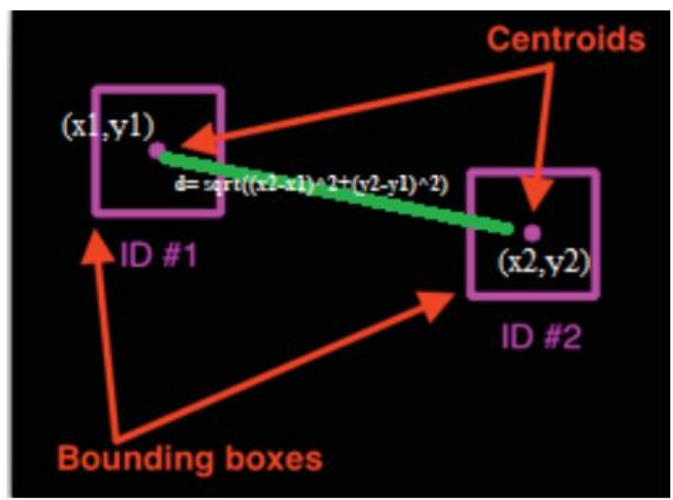

Distance Measure Between two detected boxes

3. Later finding distance, the any two adjacent person's distance which are less than 'x' pixels ('x'represents threshold in the project) distance apart from each is counted as a violation pair. Those will be highlighted in red bounding boxes and the number of violations at that particular frame will be displayed. The pair which doesn't violate will put as non-violation pair and they will be shown in green color.

$$c = \begin{cases} red & d < t \\ green & d \geq t \end{cases}$$

4. Repeat the process till the last frame of the video, and we make the count on no.of. violations occurred totally and display them as the output.

5. After Step 4, We store the no. of. Violations into local database for the flask's reference. So, when the dashboard is used the data goes through the pipeline we created locally from the database.

## D. Flowchart of the Model

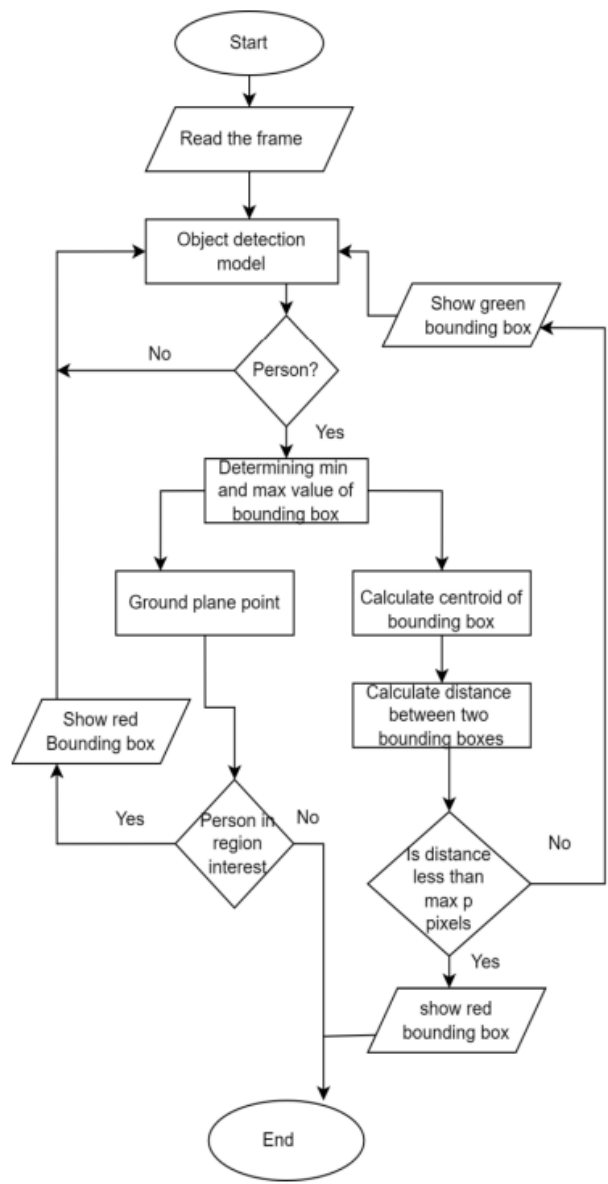

Flowchart of the model.

# V. Results and Inference

In the following segment, an extensive examination of the outcomes and their interpretation will be presented. A predetermined video showcasing people walking by is utilized as the input source. Since the camera angle in the input video remains constant, the captured footage is transformed into a 2D bird's eye view, frame by frame, enabling accurate estimation of the distances between all anomalies detected. The video's viewpoint is adjusted, and every person within the camera's scope is recognized. Each individual found in the footage is displayed by dots, and those distances, which is less than the pre-set threshold value are highlighted with red boxes.

While the detection process runs smoothly and satisfactorily within its intended range, there are occasions where errors may occur due to overlapping frames, especially when people are in close proximity to one another. To enhance the precision of distance computation, the algorithm calculates the accuracy of the distance between each individual. YOLOv3 is adept at recognizing pedestrians as objects even when only half of their bodies are visible; however, this results in less precise location matching since the bounding box extends to the lowermost edge midpoint of the half-visible body. To address the inaccuracies caused by frame overlap, a quadrilateral-shaped box is incorporated to indicate the scope of detection.

Dashboard which shows the no. of. Violations with each iterative time stamp allows the user or management to keep track of the outspread and enhance them to act accordingly. With multi-parameters it would be very beneficial for the team to track and manage in a long-haul situation.

## A. Snapshots of Results

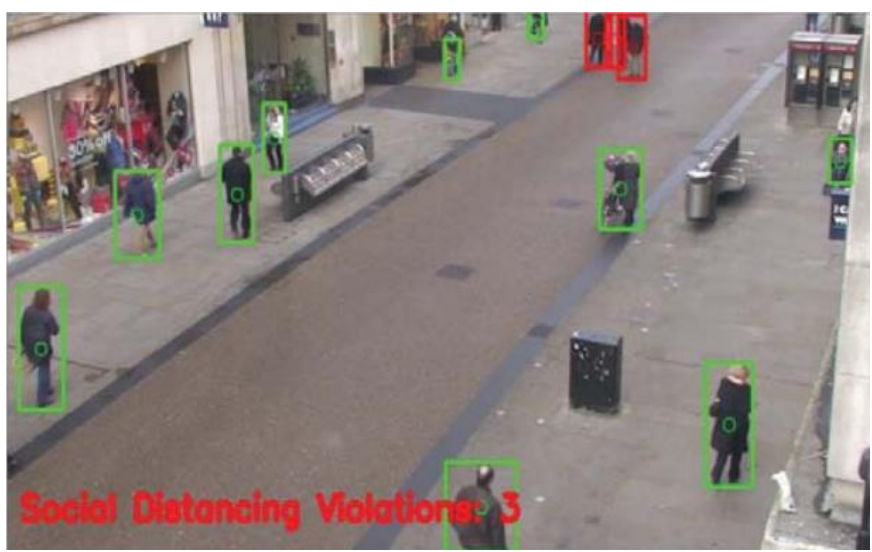

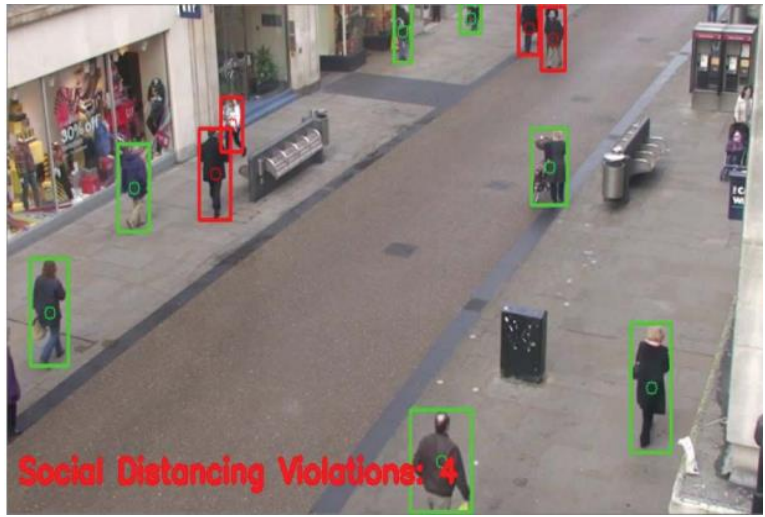

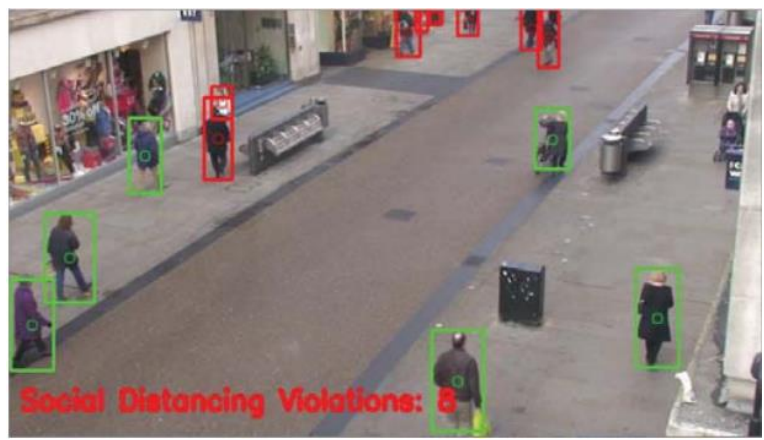

Snapshots of results highlighting the violations and with the count of violations

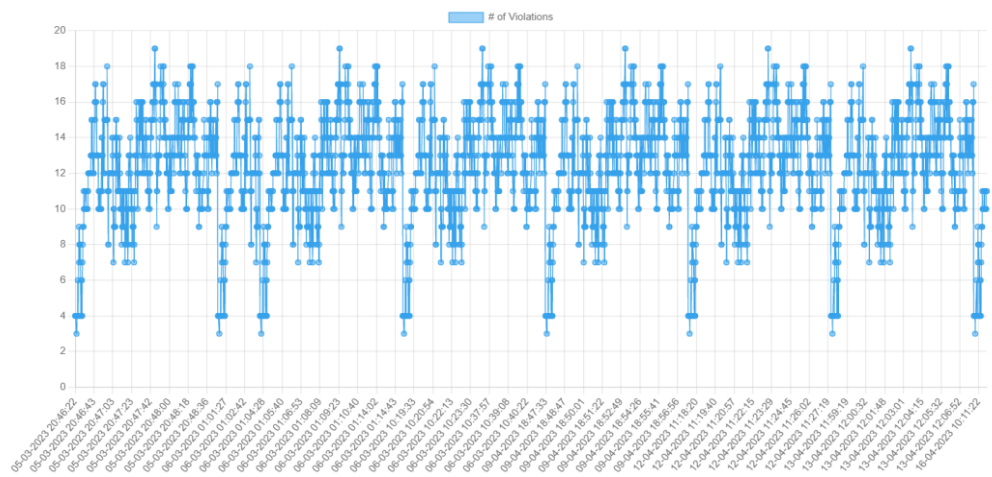

Screenshot of the dashboard created for risk Management system

## B. Factors affecting the results

Recognizing objects from images and videos is a challenging undertaking. Despite numerous studies aiming for a 99% accuracy rate, the suggested model falls short of achieving the highest levels of precision and accuracy. The factors that contribute to decreased recognition system accuracy have been identified.

1. **Weaker Resolution:** The image and videos with low resolution is not always accurate, they can be Blurry images which occur due to camera in and out focused images, interlacing of focal lengths, Low Resolution images which has low resolution spatial sensor, Acquisition mistakes which captures images with image noise.

2. **Object Size:** Though Yolov3 is capable of recognizing the object in half-form too, when the person is tiny in size, it appears as countless to the detection algorithm.

3. **Camera Calibration:** Only with proper geometrical parameters of camera is done the resolution and frames will come by steady and smooth. Improper calibration of camera can lead to very inaccurate results overall.

4. **Processing Power:** Generally, the CPUs or Laptops with higher GPU performance, works faster with better frame rates and rendition of images with more accuracy precision and accuracy. Depending upon in which kind of CPU the model is run, the results are more alike.

   The previously mentioned object detection model has been fine-tuned for binary classification purposes, distinguishing between whether or not an object is a person. This was executed on an Nvidia GTX 1060 GPU.

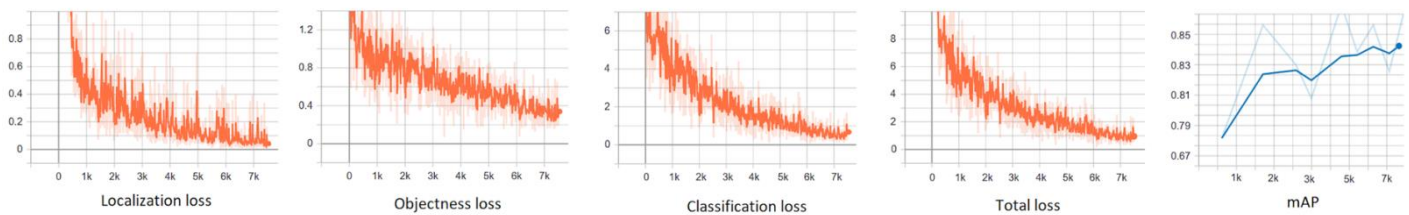

Losses per iteration of Yolov3

At final stages of the training phase, the object detection model's performance was measured based on the training time, number of iterations, mean average precision, and total loss value. YOLO v3 yielded superior results, exhibiting a balanced FPS, time for training, and F1 score. The trained

YOLO v3 model was then used to monitor social distancing in CCTV videos. A detailed differentiation study with other prevalent detection techniques, from YOLOv2, SSD, to LDCF, was conducted based on parameters such as average precision, FPS and time consumption. The comparative analysis revealed that the other methods achieved relatively less APs, ranging from 38.09% to 65.00%, in object detection, while the used YOLOv3 attained a high AP of 84.07% in detecting humans. Therefore, under the utilized conditions and device's performance, the proposed YOLO algorithm can detect humans with a high precision of 63.69% AP and at a rate of 30 FPS.

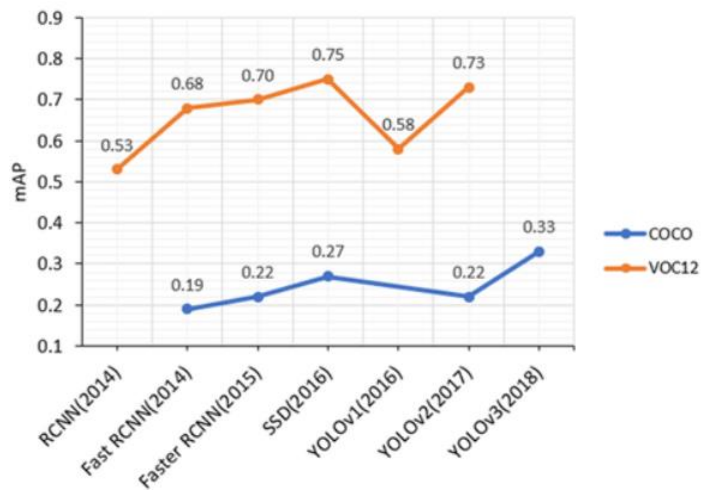

mAP of different models on COCO dataset

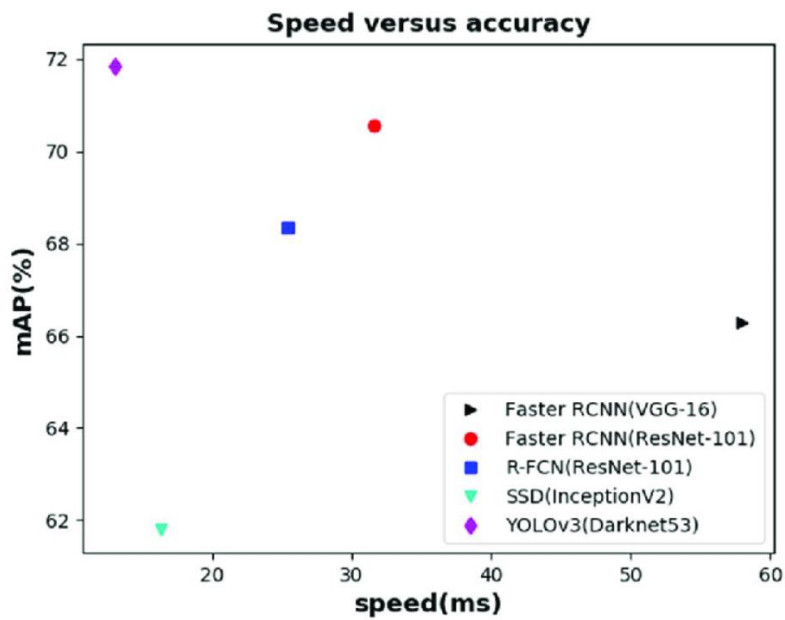

Speed vs accuracy graph of different object detection algorithms.

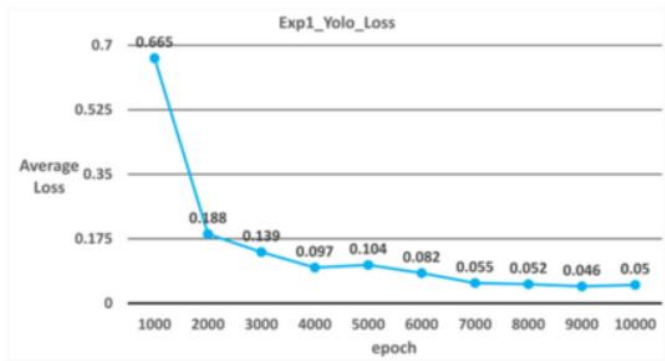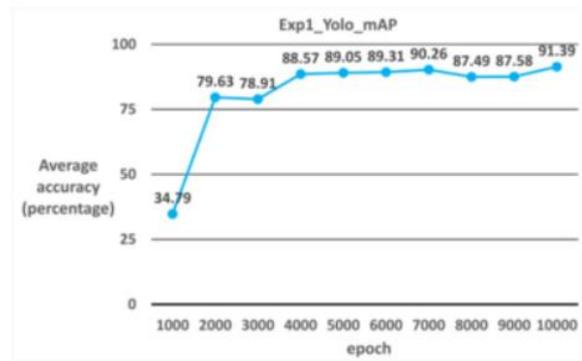

YOLOv3 avg error rate and avg. accuracy rate

## VI. Conclusion

The proposed research project addresses the issue of video social distance detection by evaluating and representing close proximity in images. Solving such challenges allows for the rapid screening of the population to detect potentially risky behaviours, particularly during pandemics. The Video Social Distance detector is not simply a computer vision problem related to mathematical proxemics, as individuals must be weighted based on the current social situation. If a group of people is found to be violating the minimum acceptable threshold value, a red bounding box is displayed. The developed model utilizes a recorded video of individuals on a public and can measure the distance between individuals. The system can differentiate and classify patterns of social distancing as "Safe" and "Unsafe" distances, as well as provide labels based on object recognition and categorization. The classifier can be used to build real-time applications and can operate with live video feeds. This device can be used with CCTV to monitor individuals during pandemics, and because mass screening is feasible, it can be employed in congested areas such as train stations, bus stops, marketplaces, streets, mall entrances, schools, colleges, and workplaces.

## VII. References

1. Degadwala, S., Vyas, D., Dave, H., & Mahajan, A. (2020). *Visual Social Distance Alert System Using Computer Vision & Deep Learning. 2020 4th International Conference on Electronics, Communication and Aerospace Technology (ICECA).*
2. Mishra, Rahul, et al. "SOCIAL DISTANCE DETECTION USING DEEP LEARNING."
3. Hou, Yew Cheong, et al. "Social distancing detection with deep learning model." *2020 8th International conference on information technology and multimedia (ICIMU)*. IEEE, 2020.
4. Punn, Narinder Singh, et al. "Monitoring COVID-19 social distancing with person detection and tracking via fine-tuned YOLO v3 and Deepsort techniques." *arXiv preprint arXiv:2005.01385* (2020).Ahmed, Imran, Misbah Ahmad, and Gwanggil Jeon. "Social distance monitoring framework using deep learning architecture to control infection transmission of COVID-19 pandemic." *Sustainable cities and society* 69 (2021): 102777.

5. Ramadass, Lalitha, Sushanth Arunachalam, and Z. Sagayasree. "Applying deep learning algorithm to maintain social distance in public place through drone technology." *International Journal of Pervasive Computing and Communications* 16.3 (2020): 223-234.
6. Salagrama, Shailaja, et al. "Real time social distance detection using Deep Learning." *2022 International Conference on Computational Intelligence and Sustainable Engineering Solutions (CISES)*. IEEE, 2022.
7. Pooranam, N., et al. "A safety measuring tool to maintain social distancing on COVID-19 using deep learning approach." *Journal of Physics: Conference Series*. Vol. 1916. No. 1. IOP Publishing, 2021.Krishna, KS Pavan, and S. Harshita. "Social distancing and face mask detection using deep learning." *Int. Res. J. Eng. Technol. (IRJET)* (2021).
8. Thylashri, S., D. Femi, and C. Thamizh Devi. "Social Distance Monitoring Method with Deep Learning to prevent Contamination Spread of Coronavirus Disease." *2022 6th International Conference on Computing Methodologies and Communication (ICCMC)*. IEEE, 2022.
9. Sriharsha, Manthri, et al. "Social distancing detector using deep learning." *Int. J. Recent Technol. Eng.(IJRTE)* 10.5 (2022): 146-149.Militante, Sammy V., and Nanette V. Dionisio. "Deep learning implementation of facemask and physical distancing detection with alarm systems." *2020 Third International Conference on Vocational Education and Electrical Engineering (ICVEE)*. IEEE, 2020.
10. Prabakaran, N., et al. "A deep learning based social distance analyzer with person detection and Tracking Using Region based convolutional neural networks for novel coronavirus." *Journal of Mobile Multimedia* (2022): 541-560.
11. Shetty, S. Vijaya, et al. "Social distancing and face mask detection using deep learning models: A survey." *2021 Asian Conference on Innovation in Technology (ASIANCON)*. IEEE, 2021.
12. Y. C. Hou, M. Z. Baharuddin, S. Yussof and S. Dzulkifly, "Social Distancing Detection with Deep Learning Model," *2020 8th International Conference on Information Technology and Multimedia (ICIMU)*, Selangor, Malaysia, 2020, pp. 334-338, doi: 10.1109/ICIMU49871.2020.9243478.
13. Z. Shao, G. Cheng, J. Ma, Z. Wang, J. Wang and D. Li, "Real-Time and Accurate UAV Pedestrian Detection for Social Distancing Monitoring in COVID-19 Pandemic," in *IEEE Transactions on Multimedia*, vol. 24, pp. 2069-2083, 2022, doi: 10.1109/TMM.2021.3075566.
14. Shalini, G. V., et al. "Social distancing analyzer using computer vision and deep learning." *Journal of Physics: Conference Series*. Vol. 1916. No. 1. IOP Publishing, 2021.
15. Krishna, KS Pavan, and S. Harshita. "Social distancing and face mask detection using deep learning." *Int. Res. J. Eng. Technol.(IRJET)* (2021).
16. Kolpe, Rupali, et al. "Identification of Face Mask and Social Distancing using YOLO Algorithm based on Machine Learning Approach." *2022 6th International Conference on Intelligent Computing and Control Systems (ICICCS)*. IEEE, 2022.
17. Khel, Muhammad Haris Kaka, et al. "Real-time monitoring of COVID-19 SOP in public gathering using deep learning technique." *Emerging Science Journal* 5 (2021): 182-196.
18. Pandiyan, Priya. "Social distance monitoring and face mask detection using deep neural network." *MSc Internet of Things with Data Analytics Bournemouth University, United Kingdom* (2020).

19. Liu C, Guo Y, Li S, Chang F. ACF Based Region Proposal Extraction for YOLOv3 Network Towards High-Performance Cyclist Detection in High Resolution Images. *Sensors*. 2019; 19(12):2671. https://doi.org/10.3390/s19122671

20. Degadwala, Sheshang & Vyas, Dhairya & Dave, Harsh & Mahajan, Arpana. (2020). Visual Social Distance Alert System Using Computer Vision & Deep Learning. 1512-1516. 10.1109/ICECA49313.2020.9297510.

21. Haq, I.U., Du, X. & Jan, H. Implementation of smart social distancing for COVID-19 based on deep learning algorithm. *Multimed Tools Appl* **81**, 33569–33589 (2022). https://doi.org/10.1007/s11042-022-13154-x

22. https://pjreddie.com/yolo/

23. https://pyimagesearch.com/2018/11/12/yolo-object-detection-with-opencv/